%% file: neurips_2026.tex
\documentclass{article}

\usepackage[preprint]{neurips_2026}
\setcitestyle{numbers,square,comma}

\usepackage[utf8]{inputenc} 
\usepackage[T1]{fontenc}    
\usepackage{hyperref}       
\usepackage{url}            
\usepackage{booktabs}       
\usepackage{amsfonts}       
\usepackage{nicefrac}       
\usepackage{microtype}      
\usepackage{xcolor}         

\usepackage{graphicx}
\usepackage{amsmath}
\usepackage{enumitem}
\usepackage{multirow}
\usepackage{caption}
\usepackage{wrapfig}
\usepackage{placeins}
\usepackage{xcolor}
\usepackage[most]{tcolorbox}
\usepackage{algorithm}
\usepackage{algorithmic}
\usepackage{cleveref}
\hypersetup{hidelinks}

\title{TAPO: Transition-Aware Policy Optimization for LLM Agents}

\input{author_info.tex}

\begin{document}

\maketitle

\begin{abstract}
  Recently, Reinforcement Learning (RL) has emerged as a crucial paradigm for the post-training of Large Language Model (LLM) agents. However, existing methods predominantly rely on sparse task rewards for policy optimization, failing to fully exploit another class of inherently dense supervisory signals naturally present during online interaction: environmental feedback following action execution. Recent theoretical studies suggest that generalization in multi-step, goal-oriented tasks hinges on predictive knowledge of environmental consequences. Inspired by this, we propose \textit{\textbf{TAPO}: \textbf{T}ransition-\textbf{A}ware \textbf{P}olicy \textbf{O}ptimization for LLM Agents}, a unified training framework that alternates between policy optimization and transition supervision. Beyond standard RL updates, TAPO repurposes rollout data to apply action-conditioned next-observation prediction supervision on a shared backbone model. This approach enhances the model's sensitivity to environmental transition dynamics and action consequences while concurrently optimizing the policy. It serves as a computationally lightweight, plug-and-play enhancement module for existing agent RL algorithms, requiring no additional expert data, extra sampling costs, or inference-time overhead. We conduct systematic experiments on WebShop and ALFWorld, integrating foundation models of various scales with different policy optimization algorithms. Empirical results demonstrate that TAPO consistently improves task performance over pure policy optimization baselines.
\end{abstract}

\section{Introduction}

The rapid advancement of Large Language Models (LLMs)~\cite{achiam2023gpt,team2023gemini,Yang2024Qwen25TR,liu2024deepseek} has catalyzed the development of autonomous agents capable of executing complex tasks. From web navigation~\cite{furuta2023multimodal,zheng2024gpt} to embodied control~\cite{li2024embodied}, LLM-based agents are demonstrating tremendous potential in solving real-world challenges.

\begin{figure}[tb]
  \vskip 0.2in
  \begin{center}
    \centerline{\includegraphics[width=\textwidth, trim=7cm 5cm 7cm 4cm, clip]{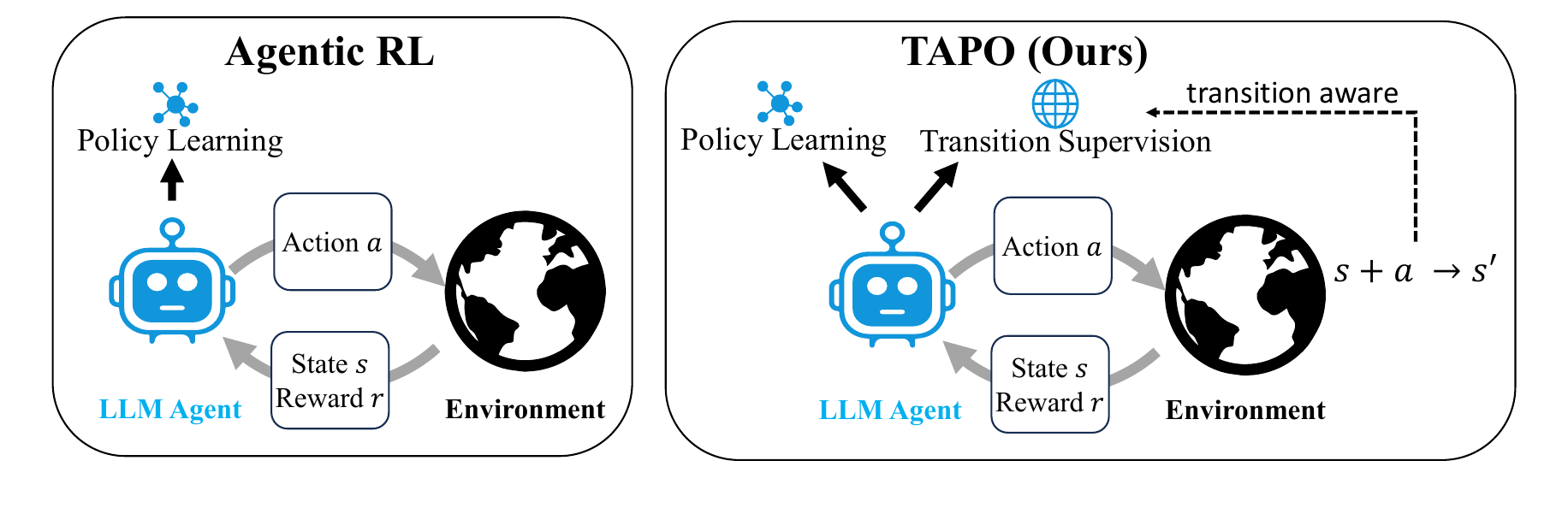}}
    \caption{
      Comparison between standard agentic RL and TAPO. Standard RL mainly learns from reward signals, whereas TAPO exploits two complementary signals from the same rollout: reward supervision for policy optimization and transition supervision for action-conditioned environment feedback modeling.
    }
    \label{fig:compare}
  \end{center}
  \vspace{-20pt}
\end{figure}

Recently, Reinforcement Learning (RL)~\cite{sutton1998reinforcement} has emerged as a crucial approach for the post-training of LLMs~\cite{ouyang2022training,jaech2024openai}. In particular, group-based RL methods like GRPO~\cite{shao2024deepseekmath} have demonstrated outstanding performance in large-scale RL training. By adopting a simple yet effective relative advantage computation, they eliminate the need for computationally expensive value models. Subsequently, methods such as RAGEN~\cite{Wang2025RAGENUS} and GiGPO~\cite{feng2025group} have extended these techniques to Agentic RL. These methods significantly enhance the multi-turn reasoning capabilities of models and optimize LLM performance in agent-based tasks. 

However, mainstream RL methods primarily rely on trial-and-error to optimize task execution. This training pipeline focuses exclusively on policy-only optimization, leaving the agent's latent understanding of environmental dynamics insufficiently constrained. The agent struggles to systematically internalize "what impact a specific action will have on the environment" during training. When confronted with long-horizon tasks characterized by sparse rewards or complex dynamics, this often leads to brittle planning trajectories and hinders performance improvements. 

Recent theoretical work~\cite{richens2025general} has demonstrated that \textit{any bounded, goal-conditioned agent capable of generalizing to multi-step goal-directed tasks must implicitly encode a predictive model of the environment within its policy}. Furthermore, prior works such as UNREAL~\cite{jaderberg2016reinforcement}, DeepMDP~\cite{gelada2019deepmdp,xu2025vlms}, and SPR~\cite{schwarzer2020data} have demonstrated that optimizing auxiliary objectives for future prediction can improve the performance of model-free RL. Inspired by this, a critical and practical question arises: During the post-training of LLM agents, how can we encourage the model to acquire this predictive knowledge of environmental consequences at the lowest possible cost? 

A natural yet frequently overlooked fact is that online RL rollouts inherently contain rich environmental supervision. For every interaction step, the agent not only receives information regarding the final task reward but also observes the actual environmental feedback following action execution. Compared to sparse reward supervision, this type of supervision is dense, localized, and action-conditioned: it explicitly tells the model "what exactly the observation will be after executing this action in the current context". However, existing RL methods do not explicitly utilize this information. 

Building on this observation, we pose a simple yet effective formulation: Is it possible to simultaneously enhance an LLM agent's policy ability and its ability to model action consequences by exclusively repurposing existing transition data from standard RL rollouts? To answer this, we propose \textit{\textbf{TAPO}: \textbf{T}ransition-\textbf{A}ware \textbf{P}olicy \textbf{O}ptimization for LLM Agents}, a unified post-training framework for LLM agents. TAPO alternates between a standard policy optimization process and a transition supervision stage. As shown in Figure~\ref{fig:compare}, the former directly optimizes task returns to improve policy capabilities; the latter trains the shared backbone model to predict the environmental feedback given the current observation and executed action, thereby heightening the model's sensitivity to environmental transition dynamics and action consequences. Because this supervisory signal is derived entirely from raw rollout data, TAPO requires no additional expert data or environment sampling, nor does it incur any extra inference overhead during testing, making it a low-cost, plug-and-play enhancement for existing agentic RL algorithms. 

Notably, while some existing works~\cite{Zhang2025AgentLV,yu2026reinforcement} have also integrated transition modeling capabilities into LLM agents, they typically treat it as a decoupled phase preceding RL, requiring extra data. In contrast, inspired by theoretical insights in reinforcement learning, we formulate transition modeling as an auxiliary objective to accelerate the policy optimization process, completely circumventing the need for extra data collection.

We systematically evaluate TAPO on two long-horizon agent benchmarks: WebShop~\cite{yao2022webshop} and ALFWorld~\cite{shridhar2020alfworld}. Empirical results demonstrate that introducing transition supervision consistently improves agent task performance across different model scales and policy optimization algorithms. These findings illustrate that, in the post-training of LLM agents, beyond task rewards, the inherent transition feedback within rollouts serves as a crucial and highly valuable learning signal in itself. 

The main contributions of this work are summarized as follows: (1) We highlight that existing agentic RL methods for LLMs primarily optimize around sparse task rewards, failing to systematically leverage action-conditioned environmental feedback. (2) We propose TAPO, a unified training framework that alternates between policy optimization and transition supervision. (3) We demonstrate consistent performance improvements across diverse environments, model scales, and RL algorithms. Further analysis reveals that transition supervision heightens the model's sensitivity to environmental feedback, thereby facilitating more effective long-horizon decision-making.

\section{Related Work}

\subsection{LLM Agents}

The emergence of LLMs has catalyzed a paradigm shift in designing autonomous agents. LLM-based agents are being extensively deployed across diverse domains, including code generation~\cite{zhang2024codeagent}, tool usage~\cite{yao2024tau}, open-ended game interaction~\cite{wang2023voyager}, web navigation~\cite{wu2025webdancer,li2025websailor}, and robotic control~\cite{zitkovich2023rt}. Early research primarily focused on prompt engineering strategies, such as ReAct~\cite{yao2022react} and Reflexion~\cite{shinn2023reflexion}, to elicit the general knowledge acquired during pre-training. While these training-free approaches avoid computational overhead, their performance is often constrained by the base model's lack of domain-specific expertise and its inability to adapt to complex environments over time.

\subsection{Reinforcement Learning for LLM Agents}

Reinforcement learning (RL) plays a pivotal role in the post-training of LLMs. Early works, such as RLHF~\cite{ouyang2022training,ziegler2019fine}, commonly adopted algorithms like PPO~\cite{schulman2017proximal} to align model outputs with human preferences. More recently, critic-free, group-based RL methods such as GRPO~\cite{shao2024deepseekmath} have gained prominence due to their computational efficiency. Methods in this family, including DAPO~\cite{yu2025dapo} and Dr.GRPO~\cite{liu2025understanding}, have demonstrated strong performance on single-turn tasks such as mathematical reasoning~\cite{guo2025deepseek}.

Recent work has extended these RL paradigms to long-horizon, multi-turn agent settings. RAGEN~\cite{Wang2025RAGENUS} treats entire interaction trajectories as training samples and uses instance filtering and gradient shaping to stabilize multi-turn learning. GiGPO~\cite{feng2025group} augments trajectory-level advantages with step-level advantages to improve credit assignment under sparse rewards. ARPO~\cite{dong2025agentic} further encourages exploration by branching rollouts when post-tool-use uncertainty is high. While these methods improve long-horizon policy optimization, they still primarily rely on scalar reward signals and often do not explicitly exploit the action-conditioned transition feedback naturally available in online interaction trajectories.

\subsection{Transition Modeling and World Models}

Modeling post-action environmental feedback has long been important in sequential decision making. In classical model-based RL~\cite{hafner2019dream,hafner2020mastering,hafner2023mastering} and recent LLM-agent work~\cite{hao2023reasoning,chae2024web,gu2024your,fang2025webevolver}, explicit world models or transition models are often learned to support planning, simulation, or improved long-horizon decision making. Recent concurrent work such as Early experience~\cite{Zhang2025AgentLV} and RWML~\cite{yu2026reinforcement} are also closely related to our setting. However, they study a different design point: they treat world-model learning itself as a stage prior to downstream policy RL, requiring either expensive expert data or additional data collection via the model.

In contrast, TAPO does not train an independent environment model or introduce a separate pre-RL stage. Instead, it studies whether rollout-derived transition supervision can be directly interleaved with standard policy optimization as a lightweight post-training recipe, focusing on reusing transition signals already present in standard RL rollouts to enhance policy optimization within a single interleaved training process.

\section{Preliminaries}

\subsection{Problem Definition}

Different from single-turn reasoning tasks, long-horizon agentic tasks require an agent to engage in multi-turn interactions with an environment based on a given task description $x \sim p(X)$. Specifically, at each time step $t$, the agent observes a state $s_t$. The LLM-based agent $\pi_\theta$ (parameterized by $\theta$), generates an action $a_t \sim \pi_\theta(\cdot | s_t, x) \in \mathcal{V}^n$ for the current step based on the task description $x$ and the observed state $s_t$, where $\mathcal{V}$ is the vocabulary of the LLM and $n$ is the maximum generation length. The environment $\mathcal{E}$ then transitions to the next state $s_{t+1}$ based on the state $s_t$ and the action $a_t$ , yielding a reward $r_t \in \mathbb{R}$.

A sequence of interactions forms a trajectory $\tau = \{(s_1, a_1, r_1), (s_2, a_2, r_2), \dots, (s_T, a_T, r_T)\}$, constituting an episode. Typically, in agentic tasks, rewards are sparse, only providing success and failure signals at the end of an episode.

The goal of agentic RL is to maximize the expected cumulative return by optimizing the policy $\pi_\theta$.

\subsection{Group-based RL for LLM agent} \label{sec:rl}

Group-based RL algorithms are widely used in the LLM agent domain as they eliminate the need for a separate value model, substantially saving storage and computational resources. Among them, GRPO is one of the most representative approaches. In these methods, the LLM agent samples $N$ trajectories $\{\tau_1, \tau_2, \dots, \tau_N\}$ for the same task with identical initial states, forming a trajectory-level group. Each trajectory $\tau_i = \{(s_1^i, a_1^i, r_1^i), (s_2^i, a_2^i, r_2^i), \dots, (s_T^i, a_T^i, r_T^i)\}$ consists of $T$ steps. The reward of trajectory $\tau_i$ is $R(\tau_i) = \sum_t r_t^i$. Within a trajectory-level group, the trajectory advantage is calculated based on the reward of each trajectory:
\begin{equation}
    A(\tau_i) = \frac{R(\tau_i) - \operatorname{mean}(\{R(\tau_j)\}_{j=1}^N)}{\operatorname{std}(\{R(\tau_j)\}_{j=1}^N)},
\end{equation}

Recent works~\cite{feng2025group,zhang2025rlvmr} usually treat each step in trajectory as an independent sample during training. The trajectory-level advantage for an action taken at any step equals the advantage of the trajectory to which it belongs:
\begin{equation}
    A_{traj}(a_t^i) = A(\tau_i), \quad \forall t \in \{1, \dots, T\}.
\end{equation}

While trajectory-level advantage provides a macroscopic signal spanning the entire trajectory, it fails to distinguish the individual contribution of actions at each specific step. To address this, GiGPO introduces a step-level advantage in addition to the trajectory-level advantage for each step, balancing global and local preferences in sparse reward scenarios. GiGPO aggregates all steps with the same environmental observation within a trajectory-level group into a step-level group and then computes the step-level advantage within this group:
\begin{equation}
    A_{step}(a_t^i) = \frac{R_t^i - \operatorname{mean} \left( \left\{ R_t^j | (s_t^j, a_t^j, R_t^j) \in \text{step\_group}_m \right\} \right)}{\operatorname{std} \left( \left\{ R_t^j | (s_t^j, a_t^j, R_t^j) \in \text{step\_group}_m \right\} \right)},
\end{equation}
where $R_t^i = \sum_{k=t}^T \gamma ^{k-t} r_k^i$, and $\gamma \in (0, 1]$ is the discount factor.

The final advantage for an action at each step is defined as:
\begin{equation}
    A(a_t^i) = \begin{cases} A_{traj}(a_t^i), & \text{GRPO} \\A_{traj}(a_t^i) + \omega \cdot A_{step}(a_t^i), & \text{GiGPO}\end{cases}
    \label{eq:adv}
\end{equation}

where $\omega \in \mathbb{R}_{\ge 0}$ is a weighting coefficient in GiGPO used to balance trajectory-level and step-level advantages.

Then RL updates the policy $\pi_{\theta}$ by maximizing the following objective:
\begin{equation}
    \mathcal{J}_{\text{RL}}(\theta) = \mathbb{E}_{\substack{x \sim p(x) \\ \tau \sim \pi_{\theta_\text{old}}}} \Bigg[\frac{1}{NT} \sum_{i=1}^N \sum_{t=1}^{T} \min \Big( r_\theta(a_t^i) A(a_t^i), \operatorname{clip}(r_\theta(a_t^i), 1 \pm \epsilon) A(a_t^i) \Big) - \beta \mathbb{D}_\text{KL}(\pi_\theta || \pi_{\text{ref}}) \Bigg],
\label{eq:rl}
\end{equation}
where $r_\theta(a_t^i) = \frac{\pi_\theta(a_t^i \mid x, s_t^i)}{\pi_{\theta_{\text{old}}}(a_t^i \mid x, s_t^i)}$ is the importance sampling ratio.

\section{Methods}

Although group-based RL furnishes an efficient policy optimization framework for LLM agents, its training signals are predominantly derived from task rewards, failing to explicitly exploit the action-conditioned transition signal naturally available in online rollouts. To address this, we propose \textit{\textbf{TAPO}}, which alternates between policy optimization and transition supervision. Specifically, TAPO reuses rollout-derived transition tuples to train the shared backbone with next-observation prediction, thereby improving both policy learning and environmental feedback modeling without extra interaction, expert data, or inference-time overhead.

\subsection{Rollout-Derived Transition Supervision}

\subsubsection{Construction of Transition Signals}

Consider a long-horizon agentic task. Conditioned on a given task description $x$, the agent interacts with the environment to yield a trajectory: $\tau=\{(s_1,a_1,r_1),(s_2,a_2,r_2),\dots,(s_T,a_T,r_T)\}.$ For each step within the trajectory, aside from the reward, we can directly observe the environmental transition: $(s_t,a_t)\rightarrow s_{t+1}.$ 

Therefore, a standard RL rollout naturally generates a set of action-conditioned transition samples. We organize these samples into a transition supervision dataset: $\mathcal{D}=\{(s_t,a_t,s_{t+1})\}.$ The crucial point here is that these triplets are not additionally collected, nor do they rely on any expert annotations; rather, they are directly sourced from the interaction trajectories already generated during the policy optimization phase. Regardless of whether a particular step ultimately belongs to a successful or a failed trajectory, its corresponding environmental feedback is veridical and can thus serve as a robust supervisory signal for learning action consequences. 

\subsubsection{Transition Supervision Objective}

Based on the aforementioned triplets, we define an action-conditioned next-observation prediction task. Specifically, given the current observation $s_t$ and the executed action $a_t$, the model is required to predict the observation $s_{t+1}$ at the next step. Denoting the model with shared parameters as $f_{\theta}$, its modeling objective can be formulated as $\hat{s}_{t+1}=f_\theta(s_t,a_t)$.

In practice, we adopt a standard teacher-forcing supervised learning approach to perform conditional likelihood modeling on the sequence of genuine next observations. Consequently, the transition supervision loss is defined as:
\begin{equation}
    \mathcal{L}_{\text{TS}}(\theta) = -\mathbb{E}_{(s_t,a_t,s_{t+1})\sim \mathcal{D}} \left[ \log p_\theta(s_{t+1}\mid s_t,a_t) \right].
    \label{eq:ts}
\end{equation}

This objective is complementary to the policy optimization objective: while the former directly optimizes task returns, the latter encourages the model to learn the structural environmental feedback following an action. Since both share the identical backbone parameters $\theta$, the representations shaped by transition supervision can directly feed back into the policy generation process, thereby enhancing the model's sensitivity to action consequences and its capacity to model environmental feedback. 

\subsection{Alternating Policy Optimization and Transition Supervision}

\begin{figure*}[tb]
  \vskip 0.2in
  \begin{center}
    \centerline{\includegraphics[width=\textwidth, trim=0cm 1.1cm 0cm 1.4cm, clip]{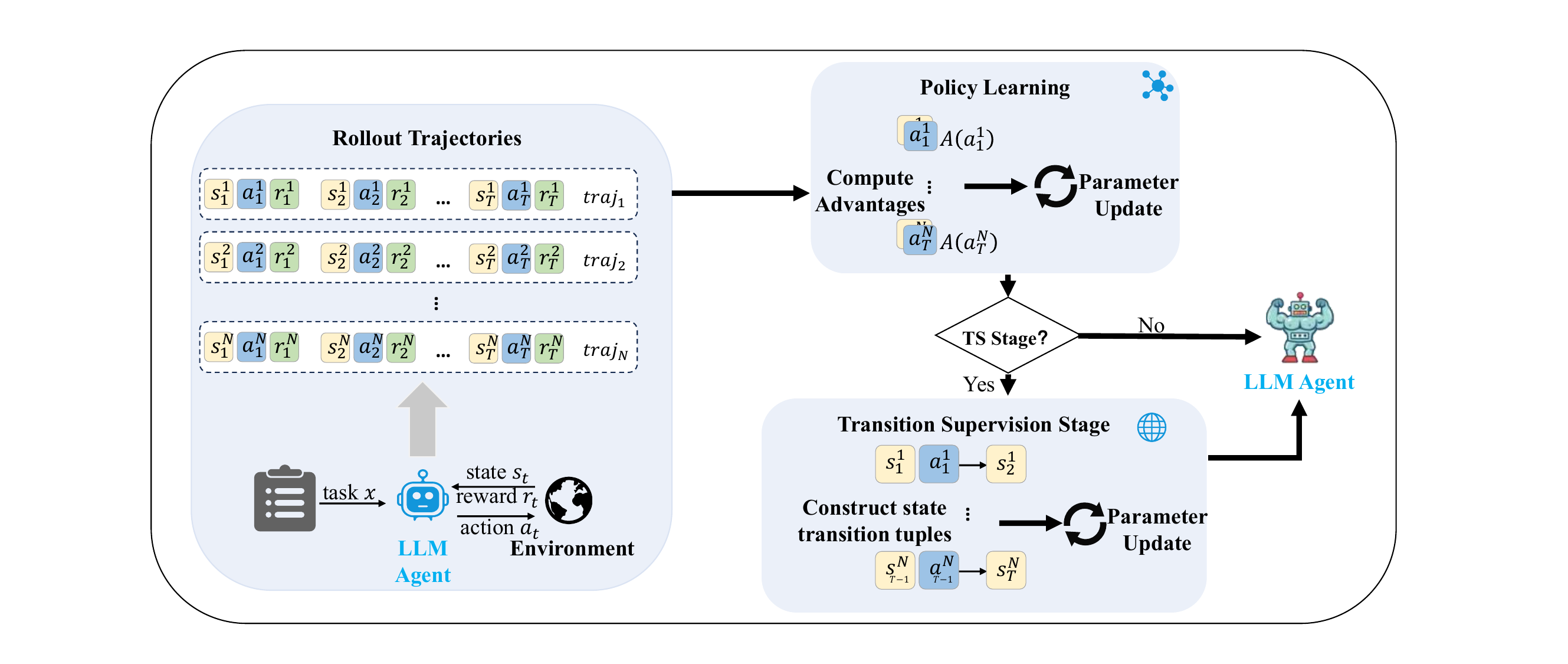}}
    \caption{
      Overview of TAPO. The LLM agent first interacts with the environment based on the task to acquire trajectories. Subsequently, these data are utilized for both policy learning and transition supervision learning. 
    }
    \label{fig:method}
  \end{center}
  \vspace{-20pt}
\end{figure*}

To simultaneously harness both reward signals and transition signals, as shown in Figure~\ref{fig:method}, we adopt a straightforward alternating training mechanism. To balance the progression of the two distinct optimization objectives and prevent either from dominating the training process, we introduce a hyperparameter $I$ to govern the alternation frequency. During training, the model first proceeds with rollouts and advantage computation following the standard agentic RL pipeline, and updates parameters based on $\mathcal{J}_{\text{RL}}(\theta)$:

\begin{equation}
\theta_{t} = \theta_t + \eta \nabla_{\theta} \mathcal{J}_{\mathrm{RL}}(\theta_t),
\end{equation}

Subsequently, at every fixed update interval $I$, we execute an additional transition supervision update: 

\begin{equation}
\theta_{t} = \theta_{t} - \eta \nabla_{\theta} \mathcal{L}_{\mathrm{TS}}(\theta_t), \quad \text{if} \quad t \bmod I = 0.
\end{equation}

In this manner, the training process periodically alternates between "learning a superior policy" and "learning action consequences". The alternation interval $I$ dictates the relative frequency between the two categories of training signals: a smaller $I$ introduces transition supervision more frequently, whereas a larger $I$ exhibits a stronger bias towards pure policy optimization. 

\section{Experiments}


    
    


\subsection{Experiment Setup}

\textbf{Benchmarks.} We trained LLM agents on two challenging benchmarks: WebShop~\cite{yao2022webshop} and ALFWorld~\cite{shridhar2020alfworld}. WebShop simulates online shopping tasks on e-commerce websites, requiring agents to comprehend human-provided textual instructions, perform multi-step web interactions, and ultimately purchase products that meet the specified criteria. ALFWorld is an embodied environment designed to evaluate an agent's ability to handle multi-step decision-making tasks in simulated household settings.

\textbf{Baselines.} We compared our method against a wide range of strong baselines, categorized as follows: (1) Closed-source LLMs: GPT-4o~\cite{achiam2023gpt} and Gemini-2.5-Pro~\cite{team2023gemini}; (2) Prompting methods: ReAct~\cite{yao2022react} and Reflexion~\cite{shinn2023reflexion}; (3) RL training methods: GRPO~\cite{shao2024deepseekmath} and GiGPO~\cite{feng2025group}. For our main experiments, we employ Qwen2.5-1.5B-Instruct and Qwen2.5-7B-Instruct as the foundation models and evaluate TAPO atop two policy optimization algorithms: GRPO and GiGPO.

\textbf{Training Details.} Unless otherwise specified, we default to the training configurations of the original papers to ensure fair comparability. In the main experiments, the alternation interval is set to $I=4$ by default (further analyzed in Section \ref{sec:afa}). For WebShop and ALFWorld, we preserve the original prompt designs, rollout group sizes, temperatures, KL penalty coefficients, and training step settings. In terms of implementation, TAPO inserts a transition supervision stage based on $(s_t, a_t, s_{t+1})$ triplets every $I$ standard RL updates. During action generation, the LLM agent is instructed to first generate a chain-of-thought enclosed within <think> </think> tags, followed by the final action enclosed within <action> </action> tags. During transition supervised learning, we prompt the model to output its prediction within <prediction> </prediction> tags, conditioned on the current state and the selected action. An anonymized code repository for reproducing the main experiments is available at \url{https://anonymous.4open.science/r/tapo-neurips2026}.

\subsection{Experiment Result}

\begin{table*}[t]
  \caption{Performance on WebShop and ALFWorld. For WebShop, we report the average score and the average success rate (\%). For ALFWorld, we report the average success rate(\%).}
  \label{tab:res}
  \begin{center}
    \begin{small}
        \setlength{\tabcolsep}{4pt}
        \begin{tabular}{lllccc}
          \toprule
          & & & \multicolumn{2}{c}{WebShop} & ALFWorld \\
          \cmidrule(l){4-6}
          Model & Type & Method & Scores & Success Rates & Success Rates\\
          \midrule
          
          GPT-4o & Prompting & GPT-4o & 31.8 & 23.7 & 48.0 \\
          Gemini-2.5-Pro & Prompting & Gemini-2.5-Pro & 42.5 & 35.9 & 60.3 \\
          \midrule
          
          \multirow{7}{*}{Qwen2.5-1.5B-Instruct} 
            & \multirow{3}{*}{Prompting} 
              & Qwen2.5-1.5B-Instruct & 23.0 & 5.2 & 4.1 \\
            & & ReAct & 40.1 & 11.3 & 12.8 \\
            & & Reflexion & 55.8 & 21.9 & 21.8 \\
            \cmidrule(l){2-6}
            & \multirow{4}{*}{RL Training} 
              & GRPO & 75.8\scriptsize{$\pm$3.5} & 56.8\scriptsize{$\pm$3.8} & 72.8\scriptsize{$\pm$3.6} \\
            & & \textbf{TAPO-GRPO(ours)} & \textbf{81.2\scriptsize{$\pm$2.8}} & \textbf{66.2\scriptsize{$\pm$4.9}} & \textbf{76.4\scriptsize{$\pm$4.0}} \\
            \cmidrule(l){3-6}
            & & GiGPO & 83.5\scriptsize{$\pm$1.8} & 67.4\scriptsize{$\pm$4.5} & 86.7\scriptsize{$\pm$1.7} \\
            & & \textbf{TAPO-GiGPO(ours)} & \textbf{86.5\scriptsize{$\pm$2.0}} & \textbf{71.7\scriptsize{$\pm$3.3}} & \textbf{88.4\scriptsize{$\pm$3.8}} \\
          \midrule
          
          \multirow{7}{*}{Qwen2.5-7B-Instruct} 
            & \multirow{3}{*}{Prompting} 
              & Qwen2.5-7B-Instruct & 26.4 & 7.8 & 14.8 \\
            & & ReAct & 46.2 & 19.5 & 31.2 \\
            & & Reflexion & 58.1 & 28.8 & 42.7 \\
            \cmidrule(l){2-6}
            & \multirow{4}{*}{RL Training} 
              & GRPO & 79.3\scriptsize{$\pm$2.8} & 66.1\scriptsize{$\pm$3.7} & 77.6\scriptsize{$\pm$5.2} \\
            & & \textbf{TAPO-GRPO(ours)} & \textbf{85.6\scriptsize{$\pm$3.9}} & \textbf{73.7\scriptsize{$\pm$4.0}} & \textbf{83.6\scriptsize{$\pm$3.9}} \\
            \cmidrule(l){3-6}
            & & GiGPO & 86.2\scriptsize{$\pm$2.6} & 75.2\scriptsize{$\pm$3.8} & 90.8\scriptsize{$\pm$1.3} \\
            & & \textbf{TAPO-GiGPO(ours)} & \textbf{89.0\scriptsize{$\pm$3.2}} & \textbf{77.9\scriptsize{$\pm$3.7}} & \textbf{93.6\scriptsize{$\pm$1.4}} \\
          \bottomrule
        \end{tabular}
    \end{small}
  \end{center}
  \vskip -0.1in
\end{table*}

Table~\ref{tab:res} presents the main results on WebShop and ALFWorld. Overall, TAPO yields consistent gains across all combinations of environments, model scales, and RL algorithms. The empirical results reveal several key observations:

First, RL fine-tuning significantly outperforms prompting baselines. All RL methods demonstrate a substantial advantage in success rates over conventional prompting techniques (e.g., ReAct and Reflexion). This validates the necessity of employing RL for post-training in complex reasoning tasks.

Second, TAPO provides consistent and significant improvements across mainstream RL algorithms. On both the Qwen2.5-1.5B-Instruct and Qwen2.5-7B-Instruct models, integrating the TAPO learning objective enables the agent to achieve superior performance compared to vanilla GRPO and GiGPO algorithms:

\begin{itemize}[leftmargin=*, noitemsep, topsep=1pt, parsep=2pt]
    \item \textbf{On the WebShop task:} For the 1.5B model, TAPO dramatically elevates the success rate of GRPO from 56.8\% to 66.2\%, and increases the average score from 75.8 to 81.2. Similarly, it further pushes the success rate of GiGPO to 71.7\%. For the 7B model, TAPO improves the success rate of GRPO training from 66.1\% to 73.7\%, while its integration with GiGPO yields a success rate increase from 75.2\% to 77.9\%.
    
    \item \textbf{On the ALFWorld task:} This trend persists. For the 1.5B model, TAPO enhances the success rate of GRPO from 72.8\% to 76.4\% and boosts GiGPO's performance to 88.4\%. For the 7B model, TAPO improves the success rates of GRPO and GiGPO by 6.0\% and 2.8\%, respectively.
\end{itemize}

\subsection{Comparison to Related Transition/World-Model Methods}

\begin{wraptable}{R}{0.55\textwidth} 
  \vspace{-15pt} 
  \centering
  \small
  \setlength{\tabcolsep}{6pt}
  \caption{Comparison with closely related transition-/world-model-based training methods on ALFWorld using Qwen2.5-7B-Instruct.}
  \label{tab:related_methods}
  \begin{tabular}{lc}
  \toprule
  Method & Success Rate (\%) \\
  \midrule
  Early Experience(IWM) & 82.8 \\
  RWML & 90.1 \\
  TAPO(ours) & \textbf{93.6} \\
  \bottomrule
  \end{tabular}
  \vspace{-5pt} 
\end{wraptable}

Since TAPO is closely related to recent methods that also exploit transition- or world-model-related signals for agent training, we additionally provide a limited reference comparison on ALFWorld with Qwen2.5-7B-Instruct. As shown in Table~\ref{tab:related_methods}, TAPO achieves a success rate of 93.6\%, compared with 82.8\% reported by Early Experience~\cite{Zhang2025AgentLV} and 90.1\% reported by RWML~\cite{yu2026reinforcement}. The result suggests that TAPO is competitive with recent closely related transition- or world-model-inspired methods.\footnote{Results for Early Experience and RWML are taken from their original papers and are included as reference only. Because these methods differ in optimization objectives, training schedules, and overall training protocols, this should not be interpreted as a fully unified head-to-head evaluation. We report these results solely to demonstrate that our method achieves competitive performance compared to state-of-the-art baselines.}

\subsection{Ablation Study}

\begin{wrapfigure}{R}{0.55\textwidth} 
  \vspace{-10pt} 
  \centering
  \includegraphics[width=\linewidth, trim=0cm 0cm 0cm 0cm, clip]{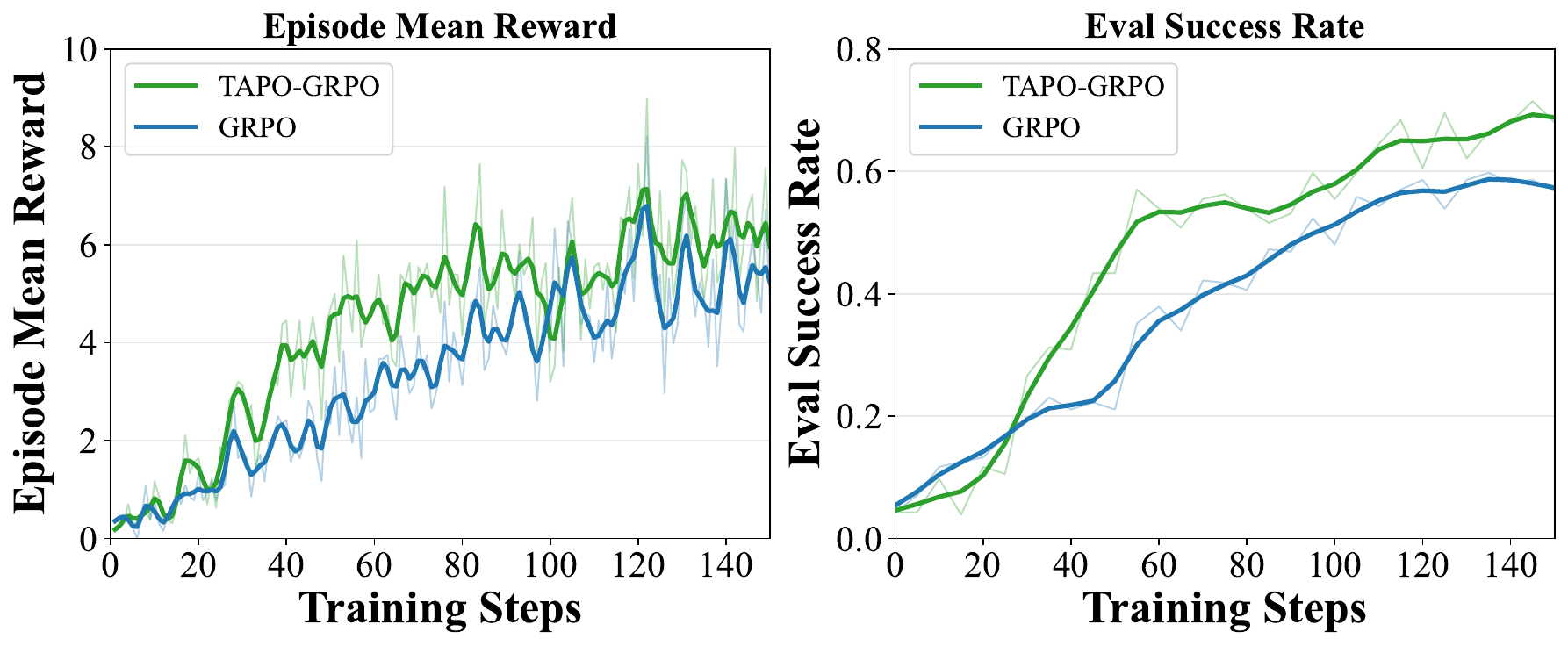}
  \caption{
    Ablation study of training dynamics on WebShop. 
  }
  \label{fig:ablation}
  \vspace{-10pt} 
\end{wrapfigure}

To isolate the contribution of TAPO, we compare vanilla GRPO and TAPO-GRPO on WebShop using Qwen2.5-1.5B-Instruct. As shown in Figure~\ref{fig:ablation}, TAPO-GRPO consistently outperforms GRPO throughout training and converges to a higher final success rate (about 66\% vs.\ 55--60\%). Moreover, its reward curve rises more steeply in the early stage, indicating improved sample efficiency under sparse rewards. 

It demonstrates that the efficacy of TAPO stems from explicitly training the agent to internalize "action-state" causal relationships, thereby facilitating more efficient and robust policy optimization in long-horizon tasks.

\subsection{In-depth Analysis} 

\subsubsection{Alternation Frequency Analysis} \label{sec:afa}

\begin{wraptable}{R}{0.45\textwidth} 
  \vspace{-10pt}
  \centering
  \small
  \setlength{\tabcolsep}{4pt} 
  \caption{Sensitivity analysis of the interleaving interval $I$ on WebShop using Qwen2.5-1.5B-Instruct with GRPO. We report success rate (SR, \%).}
  \label{tab:interval_analysis}
  \begin{tabular}{lcc}
  \toprule
  Method & $I$ & SR (\%) \\
  \midrule
  Vanilla GRPO & -- & 56.8 \\
  \midrule
  \multirow{6}{*}{TAPO-GRPO} & 2  & 63.6 \\
                               & 3  & 63.3 \\
                               & 4  & \textbf{66.2} \\
                               & 5  & 64.8 \\
                               & 10 & 62.1 \\
                               & 20 & 59.6 \\
  \bottomrule
  \end{tabular}
  \vspace{-10pt}
\end{wraptable}

Within the TAPO framework, the alternation interval $I$ serves as a critical hyperparameter controlling the balance between policy learning and transition supervision, effectively determining the relative weight of the auxiliary task within the overarching training objective. To explore the optimal configuration for this parameter, we conduct a sensitivity analysis on the WebShop benchmark using the Qwen2.5-1.5B-Instruct model combined with the GRPO algorithm, evaluating configurations where $I \in \{2, 3, 4, 5, 10, 20\}$. 

As shown in Table~\ref{tab:interval_analysis}, the agent achieves peak performance at $I=4$ with a success rate of 66.2\%, and the success rates across all other configurations remain higher than that of vanilla GRPO. This result illustrates that TAPO outperforms pure GRPO across a broadly distributed frequency range and is not hyper-sensitive to extreme values of $I$. Concurrently, a moderate alternation frequency yields optimal results, indicating the necessity of maintaining an appropriate equilibrium between policy optimization and transition supervision.

\subsubsection{Necessity of Full-Process Transition Supervision}

\begin{wraptable}{R}{0.45\textwidth} 
  \vspace{-15pt} 
  \centering
  \small
  \setlength{\tabcolsep}{4pt} 
  \caption{Effect of applying transition supervision only during early training on WebShop.}
  \label{tab:warmup_only}
  \begin{tabular}{lc}
  \toprule
  Method & SR (\%) \\
  \midrule
  Vanilla GRPO & 56.8 \\
  TAPO-GRPO (first 20 steps only) & 59.1 \\
  TAPO-GRPO (first 40 steps only) & 61.8 \\
  TAPO-GRPO(full process) & \textbf{66.2} \\
  \bottomrule
  \end{tabular}
  \vspace{-10pt} 
\end{wraptable}

A natural question arises: Does transition supervision merely serve as an initial warm-up during the early stages of training? To address this, we compare three settings on WebShop: utilizing TAPO throughout the entire process, applying transition supervision only for the first 40 steps, and applying it only for the first 20 steps. The results are shown in Table~\ref{tab:warmup_only}.

Evidently, while introducing transition supervision early on provides tangible benefits, relying on it solely as a warm-up mechanism yields final performance significantly inferior to full-process alternating training. This demonstrates that the role of TAPO extends beyond merely refining initial representations; rather, it continuously provisions the policy learning process with action-conditioned environmental feedback throughout the entire post-training phase.

\subsubsection{Transition Modeling Analysis} \label{sec:tma}

\begin{wrapfigure}{R}{0.4\textwidth} 
  \vspace{-10pt} 
  \centering
  \includegraphics[width=\linewidth, trim=0cm 0cm 0cm 0cm, clip]{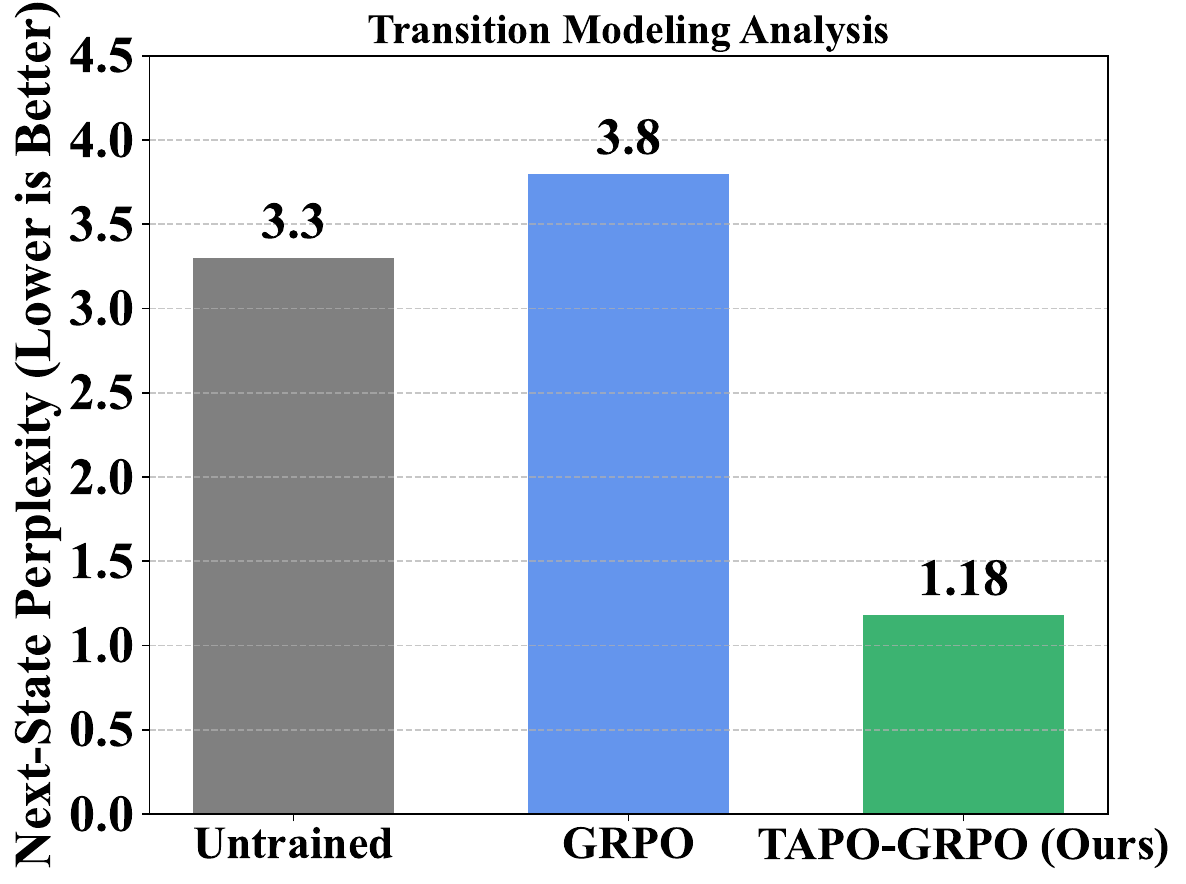}
  \caption{
    Quantitative evaluation of transition modeling. We verify the next-state prediction perplexity (PPL) on a held-out test set. 
  }
  \label{fig:ppl}
  \vspace{-15pt} 
\end{wrapfigure}

To ascertain whether the gains from TAPO align with an enhanced capacity to model action-conditioned environmental feedback, we construct an offline test set of $(s_t, a_t, s_{t+1})$ triplets from WebShop and evaluate the perplexity (PPL) of conditional next-observation modeling. We present the definition of PPL in the Appendix~\ref{appendix:ppl}.

As shown in Figure~\ref{fig:ppl}, this stark contrast indicates that vanilla GRPO fails to naturally preserve the capacity for conditional modeling of veridical environmental feedback, whereas TAPO, via explicit transition supervision, drastically enhances the model's proficiency in modeling action-conditioned environmental dynamics.

\begin{figure*}[tbp]
  \vskip 0.2in
  \begin{center}
    \centerline{\includegraphics[width=\textwidth, trim=1cm 1cm 1cm 1cm, clip]{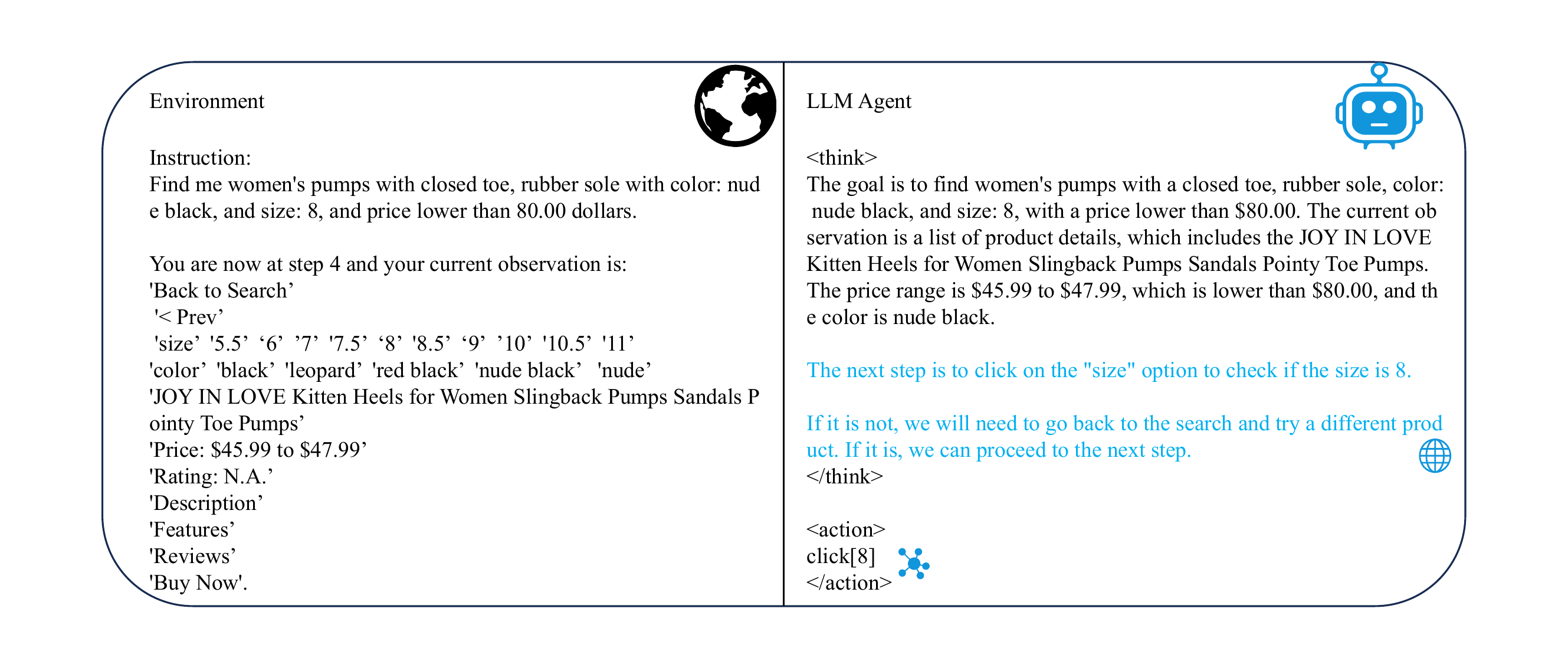}}
    \caption{Case Study of TAPO on WebShop. Instead of greedily selecting an action, the agent explicitly \textcolor{cyan}{anticipates the potential outcomes} of its action ("check if the size is 8") and \textcolor{cyan}{formulates conditional plans} ("If it is not... If it is..."). This evidence suggests that the agent has internalized the environment's transition dynamics ($s_t, a_t \to s_{t+1}$) to guide robust decision-making.}
    \label{fig:case}
  \end{center}
  \vspace{-20pt}
\end{figure*}

To investigate whether TAPO genuinely endows the agent with an understanding of environmental dynamics rather than merely relying on simplistic pattern matching, we present a detailed case study from the WebShop environment in Figure~\ref{fig:case}. During the decision-making process, the agent not only predicts the consequences of its actions (e.g., clicking an option to retrieve hidden sizing information) but also exhibits lucid conditional branching logic (i.e., formulating divergent strategies based on whether the sizing matches). This Lookahead Simulation capability provides compelling evidence that TAPO successfully enables the agent to construct an implicit understanding of environmental dynamics, transcending mere pattern matching. 

\section{Conclusion}

In this work, we propose TAPO, a post-training framework for LLM agents that alternates between policy optimization and transition supervision. It provides action-conditioned environmental feedback supervision for a shared backbone model, without necessitating extra environmental interactions, expert data, or inference-time overhead. Empirical results demonstrate that TAPO consistently improves agent performance on WebShop and ALFWorld across different model scales and policy optimization algorithms. Further analysis indicates that these gains align with an enhanced capacity for modeling action-conditioned environmental feedback, albeit accompanied by a slight trade-off in out-of-domain general capabilities (as discussed in Appendix~\ref{sec:lim}). Future work will explore more advanced supervision formulations, training scheduling, and the optimal balance between agentic task gains and the preservation of general capabilities.

\bibliographystyle{unsrt}
\bibliography{ref}

\newpage

\appendix

\section{Algorithm}

\begin{algorithm}[htbp]
\caption{TAPO}
\label{alg:tapo}
\begin{algorithmic}[1]
\REQUIRE Initial parameter $\theta$
\REQUIRE Environment $\mathcal{E}$, task distribution $p(x)$
\REQUIRE RL hyperparameters (e.g., $\omega, \epsilon, \beta$)
\REQUIRE transition learning frequency $I$, group size $N$, max step per trajectory $T$, Total training iterations $S_{Total}$
\STATE Initialize old policy $\theta_{\text{old}} \leftarrow \theta$
\FOR{iteration $s = 1$ to $S_{Total}$}
    \STATE Update the old policy model: $\theta_{\text{old}} \leftarrow \theta$
    \STATE \textbf{$\rhd$ Phase 1: Data Collection}
    \STATE Sample tasks $x \sim p(x)$.
    \STATE Reset $N$ identical environments.
    \FOR{$t=1$ to T}
        \STATE Sample actions $\{a_t^i \sim \pi_{\theta_{\text{old}}}(\cdot | s_t^i, x)\}_{i=1}^N$.
        \STATE Execute actions, get rewards $\{r_t^i\}_{i=1}^N$ and next states $\{s_{t+1}^i\}_{i=1}^N$.
    \ENDFOR
    
    \STATE \textbf{$\rhd$ Phase 2: Policy learning}
    \STATE Compute advantages $A(a_t^i)$ for each single step. (Equation \ref{eq:adv})
    \STATE Update parameters $\theta$ by maximizing objective $\mathcal{J}_{\text{RL}}(\theta)$. (Equation \ref{eq:rl})

    \STATE \textbf{$\rhd$ Phase 3: Transition Supervision Learning - Conditional}
    \IF{$s \pmod I == 0$}
        \STATE Extract all state transition triplets from current trajectories $\{\tau_i\}_{i=1}^N$ to form dataset $\mathcal{D} = \{(s_t, a_t, s_{t+1})\}$.
        \STATE Update parameters $\theta$ by minimizing loss $\mathcal{L}_{\text{TS}}(\theta)$. (Equation \ref{eq:ts})
    \ENDIF
\ENDFOR
\ENSURE Trained parameter $\theta$
\end{algorithmic}
\end{algorithm}

\section{Limitations} \label{sec:lim}

\subsection{Capability tax}
We conduct a preliminary analysis of out-of-domain general capability. On GSM8K~\cite{cobbe2021gsm8k} with Qwen2.5-1.5B-Instruct, the accuracy of the base model, GRPO, and TAPO-GRPO is 59.1, 57.9, and 56.5, respectively. These results suggest that while TAPO improves agent-task performance, it may also introduce a mild out-of-domain capability trade-off. Importantly, a similar degradation is already observed after GRPO alone, indicating that this effect is more likely a general phenomenon of task-specific post-training rather than a limitation unique to TAPO. In other words, TAPO appears to incur an additional but modest capability cost, rather than causing catastrophic collapse. Better balancing agent-task gains and general capability preservation, e.g., through improved interleaving schedules, data selection strategies, or additional regularization, is an important direction for future work.

\section{Experiment Details}

\subsection{Prompt template}

The prompt templates used in our experiments are shown in~\cref{fig:prompt_ws_pl,fig:prompt_ws_il,fig:prompt_aw_pl,fig:prompt_aw_il}. Specifically, Figures~\ref{fig:prompt_ws_pl} and ~\ref{fig:prompt_aw_pl} are utilized for policy learning, while Figures~\ref{fig:prompt_ws_il} and ~\ref{fig:prompt_aw_il} are used for Transition Supervision.

\begin{figure}[h]
\vskip 0.8in
    \centering
    \begin{tcolorbox}[
        enhanced,                 
        colback=white,            
        colframe=black,           
        coltitle=white,           
        boxrule=1pt,              
        arc=2mm,                  
        fonttitle=\bfseries,      
        title={Prompt Template for WebShop policy learning}, 
        attach boxed title to top left={xshift=0mm, yshift=0mm}, 
        halign=left,              
    ]
    You are an expert autonomous agent operating in the WebShop e-commerce environment. Your task is to: \{task\_description\}. Prior to this step, you have already taken \{step\_count\} step(s). Below are the most recent \{history\_length\} observations and the corresponding actions you took: \{action\_history\}. You are now at step \{current\_step\} and your current observation is: \{current\_observation\}. Your admissible actions of the current situation are: [\{admissible\_actions\}].
    
    \vspace{0.5em} 
    
    Now it's your turn to take one action for the current step. You should first reason step-by-step about the current situation, then think carefully which admissible action best advances the shopping goal. This reasoning process MUST be enclosed within \verb|<think>| \verb|</think>| tags. Once you've finished your reasoning, you should choose an admissible action for current step and present it within \verb|<action>| \verb|</action>| tags.
    
    \end{tcolorbox}
    \caption{The policy learning prompt template of WebShop agents.}
    \label{fig:prompt_ws_pl}
\end{figure}

\begin{figure}[h]
    \centering
    \begin{tcolorbox}[
        enhanced,                 
        colback=white,            
        colframe=black,           
        coltitle=white,           
        boxrule=1pt,              
        arc=2mm,                  
        fonttitle=\bfseries,      
        title={Prompt Template for WebShop Transition Supervision}, 
        attach boxed title to top left={xshift=0mm, yshift=0mm}, 
        halign=left,              
    ]
    You are modeling environment transitions in WebShop e-commerce Environment. Given the current textual observation and the executed action, predict the next textual observation exactly as WebShop would produce it.
    Your current observation is: \{current\_observation\}.
    Your admissible actions of the current situation are: [\{available\_actions\}]. The action you just executed is: \{action\}.

    \vspace{0.5em} 
    
    Now, based on the current observation and the executed action, predict what you will observe in the next step. Format your prediction strictly as follows:\verb|<prediction>|[Your predicted next observation in natural language, consistent with the environment]\verb|</prediction>|
    
    \end{tcolorbox}
    \caption{The Transition Supervision prompt template of WebShop agents.}
    \label{fig:prompt_ws_il}
\end{figure}

\begin{figure}[h]
    \centering
    \begin{tcolorbox}[
        enhanced,                 
        colback=white,            
        colframe=black,           
        coltitle=white,           
        boxrule=1pt,              
        arc=2mm,                  
        fonttitle=\bfseries,      
        title={Prompt Template for ALFWorld policy learning}, 
        attach boxed title to top left={xshift=0mm, yshift=0mm}, 
        halign=left,              
    ]
    You are an expert agent operating in the ALFRED embodied Environment. Your task is to: \{task\_description\}. Prior to this step, you have already taken \{step\_count\} step(s). Below are the most recent \{history\_length\} observations and the corresponding actions you took: \{action\_history\}. You are now at step \{current\_step\} and your current observation is: \{current\_observation\}. Your admissible actions of the current situation are: [\{admissible\_actions\}].
    
    \vspace{0.5em} 
    
    Now it's your turn to take an action. You should first reason step-by-step about the current situation. This reasoning process MUST be enclosed within \verb|<think>| \verb|</think>| tags. Once you've finished your reasoning, you should choose an admissible action for current step and present it within \verb|<action>| \verb|</action>| tags.
    
    \end{tcolorbox}
    \caption{The policy learning prompt template of ALFWorld agents.}
    \label{fig:prompt_aw_pl}
\end{figure}

\begin{figure}[h]
    \centering
    \begin{tcolorbox}[
        enhanced,                 
        colback=white,            
        colframe=black,           
        coltitle=white,           
        boxrule=1pt,              
        arc=2mm,                  
        fonttitle=\bfseries,      
        title={Prompt Template for ALFWorld Transition Supervision}, 
        attach boxed title to top left={xshift=0mm, yshift=0mm}, 
        halign=left,              
    ]
    You are modeling environment transitions in ALFRED Embodied Environment. Given the current textual observation and the executed action, predict the next textual observation exactly as ALFWorld would produce it.
    Your current observation is: \{current\_observation\}.
    Your admissible actions of the current situation are: [\{available\_actions\}]. The action you just executed is: \{action\}.

    \vspace{0.5em} 
    
    Now, based on the current observation and the executed action, predict what you will observe in the next step. Format your prediction strictly as follows:\verb|<prediction>|[Your predicted next observation in natural language, consistent with the environment]\verb|</prediction>|
    
    \end{tcolorbox}
    \caption{The Transition Supervision prompt template of ALFWorld agents.}
    \label{fig:prompt_aw_il}
\end{figure}

\subsection{Definition of Next-State Perplexity} \label{appendix:ppl}
To make our evaluation reproducible, we explicitly define the next-state perplexity (PPL) used in section~\ref{sec:tma}.
Given a transition tuple $(s_t, a_t, s_{t+1})$, we evaluate the likelihood of the \emph{ground-truth next-state tokens only}, conditioned on the current state $s_t$ and the executed action $a_t$.

Let the tokenized next-state sequence be
\[
s_{t+1} = (y_1, y_2, \dots, y_{L}),
\]
where $L=|s_{t+1}|$. Using teacher forcing, the conditional negative log-likelihood for one sample is
\[
\mathcal{L}_{\text{NLL}}(s_t,a_t,s_{t+1})
=
-\sum_{i=1}^{L}
\log p_\theta\!\left(y_i \mid s_t, a_t, y_{<i}\right).
\]

For a test set $\mathcal{D}_{\text{test}}=\{(s_t^{(n)},a_t^{(n)},s_{t+1}^{(n)})\}_{n=1}^{N}$, we report corpus-level next-state perplexity:
\[
\mathrm{PPL}_{\text{test}}
=
\exp\!\left(
-\frac{
\sum_{n=1}^{N}\sum_{i=1}^{L_n}
\log p_\theta\!\left(y_i^{(n)} \mid s_t^{(n)}, a_t^{(n)}, y_{<i}^{(n)}\right)
}{
\sum_{n=1}^{N} L_n
}
\right).
\]

Here, only the ground-truth tokens in the next-state sequence $s_{t+1}$ contribute to the metric; the conditioning context $(s_t,a_t)$ is excluded from the target token count. This token-level normalization ensures fair comparison across variable-length next-state observations.

\subsection{Details of Training}

We use the same training settings in ~\cite{feng2025group} for fair comparison.

\textbf{Hyperparameters for WebShop.} All methods are configured with identical hyperparameters: for policy learning, the
maximum prompt length is 4096 tokens, and the maximum response length is 512 tokens; for transition supervision learning, the maximum prompt length is 4096 tokens, and the maximum response length is 1024 tokens. Each episode is limited to 15 environment steps. The learning rate is 1e-6 for the actor. We adopt a rule-based reward, assigning a reward of 10 for success and 0 for failure. Invalid actions are penalized with a reward of -0.1. All group-based RL methods use a group size of 8 and sample 16 groups per rollout, resulting in a total of 16 × 8 = 128 environments. The rollout temperature is set to 1.0, while the validation temperature is set to 0.4. The mini-batch size is 64, and the KL-divergence loss coefficient is set to 0.01. For GiGPO, the weighting coefficient $\omega$ is set to 1 without additional tuning, and the discount factor $\gamma$ is set to 0.95.

\textbf{Hyperparameters for ALFWorld.} All methods are configured with identical hyperparameters: for policy learning, the maximum prompt length is 2048 tokens, and the maximum response length is 512 tokens; for transition supervision learning, we use the same hyperparameters. Each episode allows up to 50 environment steps. The learning rate is set to 1e-6 for the actor. We adopt a rule-based reward, assigning a reward of 10 for success and 0 for failure. To handle invalid actions generated by the agent, we apply a reward penalty of -0.1. For all group-based RL methods, we use a group size of 8 and sample 16 different groups per rollout, resulting in a total of 16 × 8 = 128 environments. The rollout temperature is set to 1.0, while the validation temperature is set to 0.4. The mini-batch size is 256, and the KL-divergence loss coefficient is set to 0.01. For GiGPO, the weighting coefficient $\omega$ is fixed at 1 without further tuning, and the discount factor $\gamma$ is set to 0.95.

\textbf{Computing Details.} For both Webshop and ALFWorld, Qwen2.5-1.5B experiments are run on 2×A100 GPUs and Qwen2.5-7B on 4×H100 GPUs, each for 150 iterations. All reported results are mean ± standard deviation over 3 independent seeds.


\end{document}

%% file: author_info.tex
\author{%
  \textbf{Cong Li$^{1}$ \quad
  Peixi Peng$^{1,2,*}$ \quad
  Yisen Zhao$^{1}$ \quad
  Xinyu Hu$^{1}$} \\
  \textbf{Shudong Liu$^{1}$ \quad
  Zhan Su$^{1}$ \quad
  Zhuojian Li$^{1}$} \\[0.75ex]
  $^{1}$School of Electronic and Computer Engineering, Peking University \\
  $^{2}$Pengcheng Laboratory \\
  $^{*}$Corresponding author: \texttt{pxpeng@pku.edu.cn}
}